\documentclass[11pt,a4paper,nohyperref]{article}
\usepackage{acl2018}
\usepackage{times}
\usepackage{latexsym}
\usepackage{balance}

\usepackage{booktabs} % For formal tables
\usepackage{times}  %Required
\usepackage{helvet}  %Required
\usepackage{courier}  %Required
\usepackage{graphicx}  %Required
\usepackage{color}
\usepackage{bbold}
\usepackage{balance}
\usepackage{booktabs} % For formal tables
\usepackage{multirow}

\usepackage{amsmath}
\usepackage{algorithmic}

\usepackage[ruled,vlined]{algorithm2e}
\usepackage{array}

\usepackage{subcaption}

\newcommand{\Ni}{({\em i})~}
\newcommand{\Nii}{({\em ii})~}

\newcommand{\Na}{{\em a})~}
\newcommand{\Nb}{{\em b})~}

\DeclareMathOperator*{\sigm}{sigm}
\DeclareMathOperator \real{\mathbb{R}}

\DeclareMathAlphabet{\pazocal}{OMS}{zplm}{m}{n}

\DeclareMathOperator*{\argmin}{\arg\!\min}

\newcommand{\Ds}{\pazocal{D}}
\newcommand{\Vs}{\pazocal{V}}

\newcommand{\Ls}{\pazocal{L}}

\newcommand{\Js}{\pazocal{J}}

\aclfinalcopy % Uncomment this line for the final submission
\def\aclpaperid{581} %  Enter the acl Paper ID here

\title{Domain Adaptation with Adversarial Training and Graph Embeddings}

\author{Firoj Alam$^*$, Shafiq Joty$
^\dagger$, \and Muhammad Imran$^*$ \\
  Qatar Computing Research Institute, HBKU, Qatar$^*$ \\
  School of Computer Science and Engineering$\dagger$\\
  Nanyang Technological University, Singapore$^\dagger$ \\
  {\tt \{fialam, mimran\}@hbku.edu.qa}$^*$ \\
  {\tt srjoty@ntu.edu.sg}$^\dagger$
  }

\date{}

\begin{document}
\maketitle

\begin{abstract}
%\vspace{-5mm}
The success of deep neural networks (DNNs) is heavily dependent on the availability of labeled data. However, obtaining labeled data is a big challenge in many real-world problems. In such scenarios, a DNN model can leverage labeled and unlabeled data from a related domain, but it has to deal with the shift in data distributions between the source and the target domains. In this paper, we study the problem of classifying social media posts during a crisis event (e.g., Earthquake). For that, we use labeled and unlabeled data from past similar events (e.g., Flood) and unlabeled data for the current event. We propose a novel model that performs adversarial learning based domain adaptation to deal with distribution drifts and graph based semi-supervised learning to leverage unlabeled data within a single unified deep learning framework. 
Our experiments with two real-world crisis datasets collected from Twitter demonstrate significant improvements over several baselines. 

\end{abstract}

%\section{Credits}

\section{Introduction}
%In recent years, deep neural networks (DNNs) have shown impressive performance gains in a wide spectrum of machine learning problems such as image recognition, language translation, speech recognition, natural language parsing,  bioinformatics, and so on. Apart from the improved performance, one crucial benefit of DNN is that they obviate the need for feature engineering and learn latent features automatically as distributed dense vectors. The success of DNNs on a problem largely depends on the availability of labeled data for that particular task. However, in many real-world application scenarios, obtaining labeled data is a big challenge and sometimes even impossible. For such problems, it may be still possible to obtain large enough labeled and unlabeled datasets from related domains. To leverage this out-of-domain data effectively, machine learning models need to deal with the drift in data distribution from the target domain data.  

The application that motivates our work is the time-critical analysis of social media (Twitter) data at the sudden-onset of an event like natural or man-made disasters~\cite{imran2015processing}. In such events, affected people post timely and useful information of various types such as reports of injured or dead people, infrastructure damage, urgent needs (e.g., food, shelter, medical assistance) on these social networks. Humanitarian organizations believe timely access to this important information from social networks can help significantly and  reduce both human loss and economic damage~\cite{P13-1159,vieweg2014integrating,power2013finding}. 

In this paper, we consider the basic task of classifying each incoming tweet during a crisis event (e.g., Earthquake) into one of the predefined classes of interest (e.g., \emph{relevant} vs. \emph{non-relevant}) in real-time. Recently, deep neural networks (DNNs) have shown great performance in classification tasks in NLP and data mining. However the success of DNNs on a task depends heavily on the availability of a large labeled dataset, which is not a feasible option in our setting (i.e., classifying tweets at the onset of an Earthquake). %In recent years there has been a growing interest in deep neural networks (DNN) and representation learning with applications to a myriad of NLP and data mining problems. 
%However, obtaining a large amount of labeled data to train an effective DNN-based classifier is infeasible.
On the other hand, in most cases, we can have access to a good amount of labeled and abundant unlabeled data from past similar events (e.g., Floods) and possibly some unlabeled data for the current event. In such situations, we need methods that can leverage the labeled and unlabeled data in a past event (we refer to this as a \emph{source} domain), and that can adapt to a new event (we refer to this as a \emph{target} domain) without requiring any labeled data in the new event. In other words, we need models that can do \emph{domain adaptation} to deal with the distribution drift between the domains and \emph{semi-supervised} learning to leverage the unlabeled data in both domains. 

%Many methods have been proposed for semi-supervised learning and domain adaptation. Existing semi-supervised methods include generative models \cite{nigam2000text}, co-training \cite{mitchell1999role}, self-training \cite{mihalcea2004co}, and graph-based models \cite{subramanya2014graph}. These methods can be categorized into two types: \emph{transductive} and \emph{inductive}. In the transductive setting, a model is only applicable to the unlabeled instances observed at training time, that is, the model does not generalize to unobserved instances. Whereas, an inductive model generalizes to data that are not seen at the training time. Therefore, it is more desirable to have an inductive model over a transductive one.  

%Domain adaption methods attempt to learn a correspondence (or mapping) between the source and the target domains so that the model learned from the source domain can also be used to classify instances in the target domain \cite{daume2009frustratingly}. This correspondence can be learned in the absence of any labeled data in the target domain or using few labeled instances. We focus on the former case, although our method can be easily generalized to the later situation. 

Most recent approaches to semi-supervised learning \cite{yang2016revisiting} and domain adaptation \cite{ganin2016domain} use the automatic feature learning capability of DNN models. In this paper, we extend these methods by proposing a novel model that performs domain adaptation and semi-supervised learning within a single unified deep learning framework. In this framework, the basic task-solving network (a convolutional neural network in our case) is put together with two other networks -- one for semi-supervised learning and the other for domain adaptation. The semi-supervised component learns internal representations (features) by predicting contextual nodes in a graph that encodes \emph{similarity} between labeled and unlabeled training instances. The domain adaptation is achieved by training the  feature extractor (or encoder) in \emph{adversary} with respect to a domain discriminator, a binary classifier that tries to distinguish the domains. %using a domain discriminator, which is a binary classifier that tries to decide whether the input example comes from the source or from the target domain. The training of this domain discriminator network is \emph{adversarial} with respect to the shared layers by using gradient reversal during backpropagation, which makes the training to \emph{maximize} the loss of the discriminator rather than to minimize it \cite{ganin2016domain}. 
The overall idea is to learn high-level abstract representation that is discriminative for the main classification task, but is invariant across the domains. We propose a stochastic gradient descent (SGD) algorithm to train the components of our model simultaneously.    

The evaluation of our proposed model is conducted using two Twitter datasets on scenarios where there is only unlabeled data in the target domain. Our results demonstrate the following.

\begin{enumerate}	\itemsep-0.2em
\item When the network combines the semi-supervised component with the supervised component, depending on the amount of labeled data used, it gives $5\%$ to $26\%$ absolute gains in F1 compared to when it uses only the supervised component.

\item Domain adaptation with adversarial training improves over the adaptation baseline (i.e., a transfer model) by $1.8\%$ to $4.1\%$ absolute F1.

\item When the network combines domain adversarial training with semi-supervised learning, we get further gains ranging from 5\% to 7\% absolute in F1 across events.
\end{enumerate}

Our source code is available on Github\footnote{https://github.com/firojalam/domain-adaptation} and the data is available on CrisisNLP\footnote{http://crisisnlp.qcri.org}.

The rest of the paper is organized as follows. In Section~\ref{sec:methodology}, we present the proposed method, i.e., domain adaptation and semi-supervised graph embedding  learning. In Section~\ref{sec:experiments}, we present the experimental setup and baselines. The results and analysis are presented in Section~\ref{sec:results}. %We also provide discussion on the results in Section~\ref{sec:experiments}. 
In Section~\ref{sec:related_works}, we present the works relevant to this study. Finally, conclusions appear in Section ~\ref{sec:conclusions}.

%\section{Methodology}
\section{The Model}
\label{sec:methodology}
\begin{figure*}[t]
	\centering
	\includegraphics[height=2.6in]{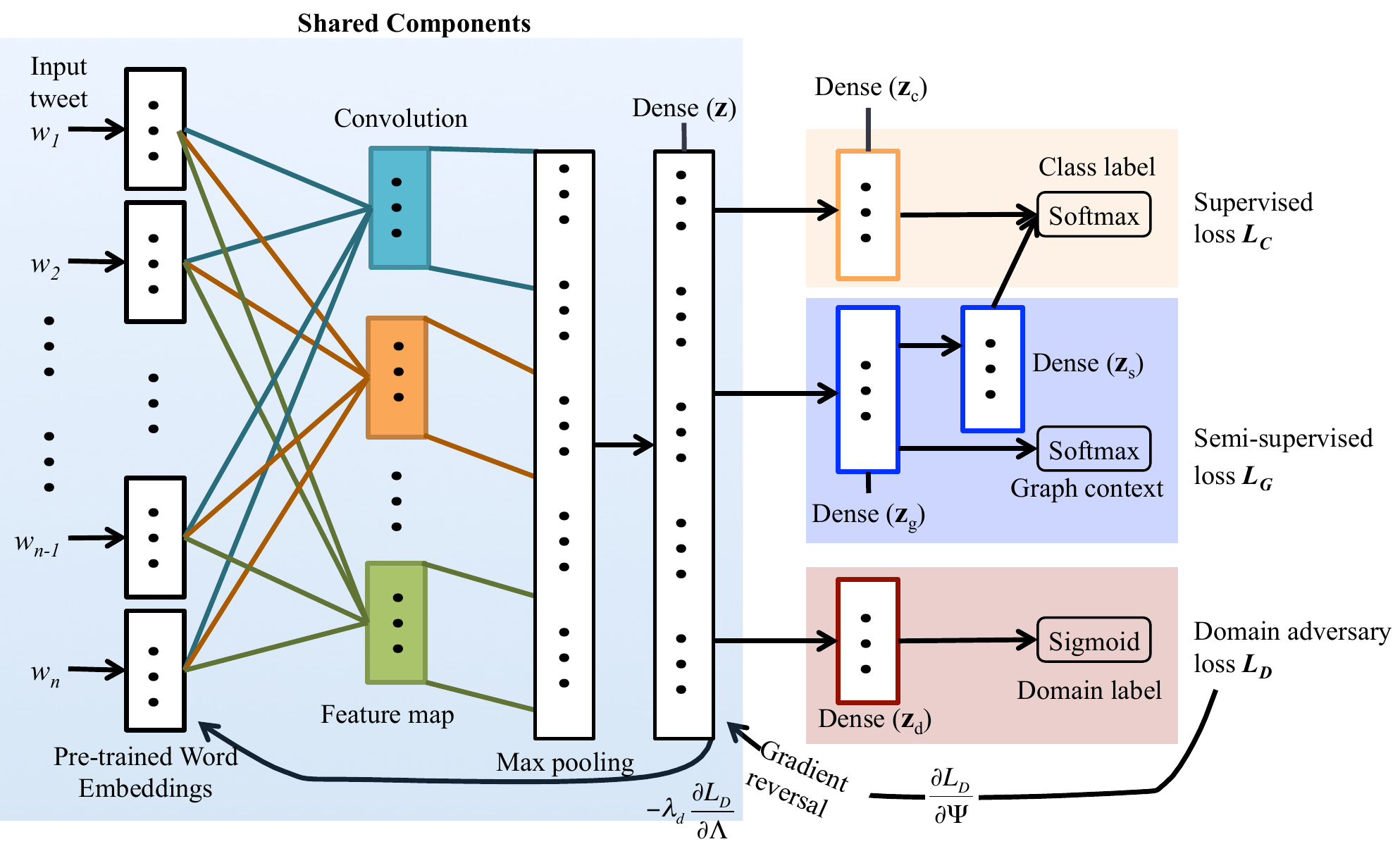}
	\caption{The system architecture of the domain adversarial network with graph-based semi-supervised learning. The shared components part is shared by supervised, semi-supervised and domain classifier. 
    }   
	\label{fig:cnn_graph}
%	\vspace{-1em}
\end{figure*}

We demonstrate our approach for domain adaptation with adversarial training and graph embedding on a tweet classification task  to support crisis response efforts. Let $\Ds_S^l = \{\mathbf{t}_i, y_{i}\}_{i=1}^{L_s}$ and $\Ds_S^u = \{\mathbf{t}_i\}_{i=1}^{U_s}$ be the set of labeled and unlabeled tweets for a source crisis event $S$ (e.g., Nepal earthquake), where $y_i \in \{1, \ldots, K\}$ is the class label for tweet $\mathbf{t}_i$, $L_s$ and $U_s$ are the number of labeled and unlabeled tweets for the source event, respectively. In addition, we have unlabeled tweets $\Ds_T^u = \{\mathbf{t}_i\}_{i=1}^{U_t}$ for a target event $T$ (e.g., Queensland flood) with ${U_t}$ being the number of unlabeled tweets in the target domain. Our ultimate goal is to train a cross-domain model $p(y|\mathbf{t}, \theta)$ with parameters $\theta$ that can classify any tweet in the target event $T$ without having any  information about class labels in $T$.

Figure \ref{fig:cnn_graph} shows the overall architecture of our neural model. The input to the network is a tweet $\mathbf{t}=(w_1, \ldots, w_n)$ containing words that come from a finite vocabulary $\Vs$ defined from the training set. The first layer of the network maps each of these words into a distributed representation $\real^d$ by looking up a shared embedding matrix $E$ $\in$ $\real^{|\Vs| \times d}$. We initialize the embedding matrix $E$ in our network with word embeddings that are pretrained on a large crisis dataset (Subsection \ref{ssec:word_embedding}). However, embedding matrix $E$ can also be initialize randomly. 
%We can initialize $E$ randomly or using pretrained word vectors like word2vec \cite{mikolov2013efficient} (more in Subsection \ref{ssec:word_embedding}). 
The output of the look-up layer is a matrix  $X \in \real^{n \times d}$, which is passed through a number of convolution and pooling layers to learn higher-level feature representations. A \emph{convolution} operation applies a \emph{filter} $\mathbf{u} \in \real^{k.d}$ to a window of $k$ vectors to produce a new feature $h_t$ as 

\vspace{-1em}
\begin{eqnarray}
h_t = f(\mathbf{u} . X_{t:t+k-1}) 
\end{eqnarray}

\noindent where $X_{t:t+k-1}$ is the concatenation of $k$ look-up vectors, and $f$ is a nonlinear activation; we use rectified linear units or ReLU. We apply this filter to each possible $k$-length windows in $X$ with stride size of $1$ to generate a \emph{feature map} $\mathbf{h}^j$ as: 

\vspace{-1em}
\begin{eqnarray}
\mathbf{h}^j = [h_1, \ldots, h_{n+k-1}]
\end{eqnarray}
%\vspace{-1.5mm}

\noindent We repeat this  process $N$ times with $N$ different filters to get $N$ different feature maps. We use a \emph{wide} convolution \cite{kalchbrenner2014convolutional}, which ensures that the filters reach the entire tweet, including the boundary words. This is done by performing \emph{zero-padding}, where out-of-range (i.e., $t$$<$$1$ or $t$$>$$n$) vectors are assumed to be zero. With wide convolution, $o$ zero-padding size and $1$ stride size, each feature map contains $(n + 2o - k + 1)$ convoluted features. After the convolution, we apply a \emph{max-pooling} operation to each of the feature maps, 

\vspace{-1em}
\begin{eqnarray}
\mathbf{m} = [\mu_p(\mathbf{h}^1), \cdots, \mu_p(\mathbf{h}^N)]
\end{eqnarray}
%\vspace{-1.5mm}

\noindent where $\mu_p(\mathbf{h}^j)$ refers to the $\max$ operation applied to each window of $p$ features with stride size of $1$ in the feature map $\mathbf{h}^i$. 
%For instance, for $p=2$, this pooling gives the same number of features as in the feature map (because of the zero-padding).
Intuitively, the convolution operation composes local features into higher-level representations in the feature maps, and max-pooling extracts the most important aspects of each feature map while reducing the output dimensionality. Since each convolution-pooling operation is performed independently, the features extracted become invariant in order (i.e., where they occur in the tweet). To incorporate order information between the pooled features, we include a fully-connected (dense) layer 

%\vspace{-0.5em}
\begin{equation}
\mathbf{z} = f(V\mathbf{m}) 
\label{dense} 
\end{equation}
%\vspace{-1.5mm}

\noindent where $V$ is the weight matrix. We choose a convolutional architecture for feature composition because it has shown impressive results on similar tasks in a supervised setting \cite{ICWSM1715655}. 

The network at this point splits into three branches (shaded with three different colors in Figure \ref{fig:cnn_graph}) each of which serves a different purpose and contributes a separate loss to the overall loss of the model as defined below: 

\vspace{-1em}
\begingroup\makeatletter\def\f@size{8}\check@mathfonts
\begin{eqnarray}
\Ls(\Lambda, \Phi, \Omega, \Psi) = \Ls_C(\Lambda, \Phi) +  \lambda_g \Ls_G(\Lambda, \Omega) + \lambda_d \Ls_D(\Lambda, \Psi) 
\label{eq:fullmodel}
\end{eqnarray}
\endgroup
%\vspace{-1.5mm}
\noindent where $\Lambda=\{U, V\}$ are the convolutional filters and dense layer weights that are shared across the three branches. The first component $\Ls_C(\Lambda, \Phi)$ is a supervised classification loss based on the labeled data in the source event. The second component $\Ls_G(\Lambda, \Omega)$ is a graph-based semi-supervised loss that utilizes both labeled and unlabeled data in the source and target events to induce structural similarity between training instances. The third component $\Ls_D(\Lambda, \Omega)$ is an adversary loss that again uses all available data in the source and target domains to induce domain invariance in the learned features. The tunable hyperparameters $\lambda_g$ and $\lambda_d$ control the relative strength of the components. %As can be noticed in Figure \ref{fig:cnn_graph} (also in Equation \ref{eq:fullmodel}) that the three components of the network share a common     
%In the following subsections, we elaborate on these components of the model as we formally define them.          

\subsection{Supervised Component} 
\label{subsec:sup}

The supervised component induces label information (e.g., \emph{relevant} vs. \emph{non-relevant}) directly in the network through the classification loss $\Ls_C(\Lambda, \Phi)$, which is computed on the labeled instances in the source event, $\Ds_S^l$. Specifically, this branch  of the network, as shown at the top in Figure \ref{fig:cnn_graph}, takes the shared representations $\mathbf{z}$ as input and pass it through a task-specific dense layer 

\vspace{-0.6em}
\begin{equation}
\mathbf{z}_c = f(V_c \mathbf{z}) \label{dense2} 
\end{equation}
%\vspace{-1mm}
\noindent where $V_c$ is the corresponding weight matrix. The activations $\mathbf{z}_c$ along with the activations from the semi-supervised branch $\mathbf{z}_s$ are used for classification. More formally, the classification layer defines a Softmax    

\vspace{-0.6em}
\begin{eqnarray}
p(y=k|\mathbf{t}, \theta) = \frac{\exp \left(W_{k}^T \left[\mathbf{z}_c; \mathbf{z}_s \right] \right)} {\sum_{k'} \exp \left( {W}_{k'}^T \left[\mathbf{z}_c; \mathbf{z}_s \right] \right)}
\end{eqnarray}
%\vspace{-1mm}
\noindent where $[.;.]$ denotes concatenation of two column vectors, ${W}_{k}$ are the class weights, and $\theta = \{ U, V, V_c, W\}$ defines the relevant parameters for this branch of the network with $\Lambda = \{U,V\}$ being the shared parameters and $\Phi = \{V_c, W\}$ being the parameters specific to this branch. Once learned, we use $\theta$ for prediction on test tweets. The classification loss $\Ls_C(\Lambda, \Phi)$ (or $\Ls_C(\theta)$) is defined as  

\vspace{-1em}
\begingroup\makeatletter\def\f@size{9}\check@mathfonts
\begin{eqnarray}
\Ls_C(\Lambda, \Phi) = {- \frac{1}{L_s} \sum_{i=1}^{L_s} \mathbb{I} (y_i = k) \log p(y_{i} = k | \mathbf{t}_{i}, \Lambda, \Phi)}
\end{eqnarray}
\endgroup
%\vspace{-1mm}

\noindent where $\mathbb{I}(.)$ is an indicator function that returns $1$ when the argument is true, otherwise it returns $0$. 

\subsection{Semi-supervised Component}  
\label{subsec:semisup}
The semi-supervised branch (shown at the middle in Figure \ref{fig:cnn_graph}) induces structural similarity between training instances (labeled or unlabeled) in the source and target events. We adopt the recently proposed graph-based semi-supervised deep learning framework \cite{yang2016revisiting}, which shows impressive gains over existing semi-supervised methods on multiple datasets. In this framework, a ``similarity'' graph $G$ first encodes relations between training instances, which is then used by the network to learn internal representations (i.e., embeddings). 
%In the interest of coherence, we first describe how our network learns embeddings from the graph context assuming that the graph is already constructed, then we describe how we build the graph.  

\subsubsection{Learning Graph Embeddings}  
\label{subsubsec:emb}
The semi-supervised branch takes the shared representation $\mathbf{z}$ as input and learns internal representations by predicting a node in the graph context of the input tweet. Following \cite{yang2016revisiting}, we use \emph{negative sampling} to compute the loss for predicting the  context node, and we sample two types of contextual nodes: \Ni one is based on the graph $G$ to encode  structural information, and \Nii the second is based on the labels in $\Ds_S^l$ to incorporate label information through this branch of the network. The ratio of positive and negative samples is controlled by a random variable $\rho_1 \in (0,1)$, and the proportion of the two context types is controlled by another random variable $\rho_2 \in (0,1)$; {see Algorithm 1 of \cite{yang2016revisiting} for details on the sampling procedure.} 

Let $( j, \gamma)$ is a tuple sampled from the distribution $p(j, \gamma|i, \Ds_S^l, G)$, where $j$ is a context node of an input node $i$ and $\gamma \in \{+1,-1\}$ denotes whether it is a positive or a negative sample; $\gamma = +1$ if $\mathbf{t}_i$ and $\mathbf{t}_j$ are neighbors in the graph (for graph-based context) or they both have same labels (for label-based context), otherwise $\gamma = -1$. The negative log loss for context prediction $\Ls_G(\Lambda, \Omega)$ can be written as 
%\vspace{-2mm}
\begingroup\makeatletter\def\f@size{8}\check@mathfonts
\begin{eqnarray}
\Ls_G(\Lambda, \Omega) = - \frac{1}{L_s+U_s} \sum_{i=1}^{L_s+U_s} \mathbb{E}_{(j,\gamma)} \log \sigma \left( \gamma {C}_{j}^{T} \mathbf{z}_g (i) \right) 
\end{eqnarray}
%\vspace{-1mm}
\endgroup
\noindent where $\mathbf{z}_g (i) = f(V_g \mathbf{z} (i))$ defines another dense layer (marked as \emph{Dense ($\mathbf{z}_g$)} in Figure \ref{fig:cnn_graph}) having weights $V_g$, and ${C}_j$ is the weight vector associated with the context node $\mathbf{t}_j$. Note that here $\Lambda = \{U,V\}$ defines the shared parameters and $\Omega = \{V_g, C\}$ defines the parameters specific to the semi-supervised branch of the network.

\subsubsection{Graph Construction} 
\label{ssec:graph-consturction}

%For graph construction, we followed similarity based graph construction approach. 
%Given a set of $n$ instances, i.e., tweets in our training set, a 
% approach is to construct the graph
Typically graphs are constructed based on a relational knowledge source, e.g., citation links in \cite{lu:icml03}, or distance between instances \cite{zhu05survey}. However, we do not have access to such a relational knowledge in our setting. On the other hand, computing distance between $n(n-1)/2$ pairs of instances to construct the graph is also very expensive \cite{muja2014scalable}. Therefore, we choose to use k-nearest neighbor-based approach as it has been successfully used in other study \cite{steinbach2000comparison}.
 %for finding nearest neighbors of instances 
% as it has been shown to be an effective approach in other studies \cite{dong2011efficient,jebara2009graph}. 

The nearest neighbor graph consists of $n$ vertices and for each vertex, there is an edge set consisting of a subset of $n$ instances, i.e., tweets in our training set. The edge is defined by the distance measure $d({i},{j})$ between tweets $\mathbf{t}_{i}$ and $\mathbf{t}_{j}$, where the value of $d$ represents how similar the two tweets are. %close or far they are to each other. %where small value of $d$ represents how close they are to each other. 
We used k-d tree data structure \cite{bentley1975multidimensional} to efficiently find the nearest instances. 
%--commented by Firoj: The reason of choosing this approach is that if $\mathbf{t}_{i}$ is very far from $\mathbf{t}_{j}$ and $\mathbf{t}_{k}$ is close to $\mathbf{t}_{j}$ then without computing the distance between $\mathbf{t}_{i}$ and $\mathbf{t}_{k}$ we can infer that they are far. 
To construct the graph, we first represent each tweet by averaging the word2vec vectors of its words, and then we measure $d(i,j)$ by computing the \textit{Euclidean} distance between the vectors. The number of nearest neighbor $k$ was set to 10. The reason of averaging the word vectors is that it is computationally simpler and it captures the relevant semantic information for our task in hand. Likewise,  
%has success in many text classification tasks \cite{zhang2015sensitivity,abdelwahab2016uofl}. 
we choose to use Euclidean distance instead of cosine  for computational efficiency.  
%%It is needed to mention that averaging word2vec vector is also well studied in many text classification tasks such as sentiment classification of tweets and  document classification \cite{zhang2015sensitivity,abdelwahab2016uofl}.

\subsection{Domain Adversarial Component}  
\label{subsec:domainadv}
The network described so far can learn abstract features through convolutional and dense layers that are discriminative for the classification task (\emph{relevant} vs. \emph{non-relevant}). The supervised branch of the network uses labels in the source event to induce label information directly, whereas the semi-supervised branch induces similarity information between labeled and unlabeled instances. However, our goal is also to make these learned features invariant across domains or events (e.g., \emph{Nepal Earthquake} vs. \emph{Queensland Flood}). We achieve this by domain \emph{adversarial} training of neural networks \cite{ganin2016domain}.

We put a domain discriminator, another branch in the network (shown at the bottom in Figure \ref{fig:cnn_graph}) that takes the shared internal representation $\mathbf{z}$ as input, and tries to discriminate between the domains of the input --- in our case, whether the input tweet is from $\Ds_S$ or from $\Ds_T$. The domain discriminator is defined by a sigmoid function: 

\vspace{-1em}
\begin{eqnarray}
\hat{\delta} = p(d = 1|\mathbf{t}, \Lambda, \Psi) = \sigm (\mathbf{w}_d^T \mathbf{z}_d)
\end{eqnarray}
%\vspace{-1mm}
%$\hat{l}_{\omega} = p(l = 1|\mathbf{f}, \omega) = \sigm (\mathbf{w}_l^T \mathbf{h}_l)$, 
\noindent where $d\in\{0,1\}$ denotes the domain of the input tweet $\mathbf{t}$, $\mathbf{w}_d$ are the final layer weights of the discriminator, and $\mathbf{z}_d = f(V_d \mathbf{z})$ defines the hidden layer of the discriminator with layer weights $V_d$. Here $\Lambda = \{U,V\}$ defines the shared parameters, and $\Psi = \{ V_d, \mathbf{w}_d\}$ defines the parameters specific to the domain discriminator. We use the negative log-probability as the discrimination loss: 

\vspace{-1em}
\begingroup\makeatletter\def\f@size{10}\check@mathfonts
\begin{eqnarray}
\Js_i(\Lambda, \Psi) = {- d_i \log \hat{\delta}  - (1-d_i) \log \left(1- \hat{\delta} \right)}
\end{eqnarray}
\endgroup

\noindent  We can write the overall domain adversary loss over the source and target domains as 
\begingroup\makeatletter\def\f@size{8.5}\check@mathfonts
\begin{eqnarray}
%\vspace{-2mm}
\Ls_D(\Lambda, \Psi) =  - \frac{1}{L_s+U_s} \hspace{-0.9mm} \sum_{i=1}^{L_s+U_s} \hspace{-0.9mm} \Js_i(\Lambda, \Psi) - \frac{1}{U_t} \sum_{i=1}^{U_t} \Js_i(\Lambda, \Psi) \label{loss}
\end{eqnarray}
\endgroup
%\vspace{-1mm}
\noindent where $L_s+U_s$ and $U_t$ are the number of training instances in the source and target domains, respectively. In adversarial training, we seek parameters (saddle point) such that   
% $\theta = \{U,V,\mathbf{w}\}$, $\omega = \{U,V,\mathbf{w},U_l,\mathbf{w}_l\}$, and the hyper-parameter $\lambda$ controls the relative strength of the two networks. 
%\vspace{-2mm}
\begin{eqnarray}
\theta^* = \argmin_{\Lambda, \Phi, \Omega} \max_{\Psi} \Ls (\Lambda,\Phi, \Omega, \Psi) 
\end{eqnarray}
%\vspace{-1mm}
\noindent which involves a maximization with respect to $\Psi$ and a minimization with respect to $\{\Lambda, \Phi, \Omega\}$. In other words, the updates of the shared parameters $\Lambda = \{U, V\}$ for the discriminator work adversarially to the rest of the network, and vice versa. This is achieved by reversing the gradients of the discrimination loss $\Ls_D(\Lambda, \Psi)$, when they are backpropagated to the shared layers (see Figure \ref{fig:cnn_graph}). 

%%  $\{U,V,\mathbf{w}\}$  two classifiers work adversarially to each other.  language  $\Ls_l(\omega)$

\subsection{Model Training}  
\label{subsec:training}
\setlength{\textfloatsep}{0.10in}
%\scalebox{0.75}{
\begin{algorithm}[t!]
\small
\SetKwInOut{Input}{Input}\SetKwInOut{Output}{Output}
%\SetAlgoLined
\SetAlgoNoLine
\SetNlSkip{0em}
\Input{data $\Ds_S^l$, $\Ds_S^u$, $\Ds_T^u$; graph $G$}
\Output{learned parameters  $\theta = \{\Lambda, \Phi \}$}
1. Initialize model parameters $\{E, \Lambda, \Phi, \Omega, \Psi \}$; \\
%	\hspace{0.5cm} a. $\psi_c$ for $\Lb_c (\mathbf{v}_i)$ \tcp*{over words}
%    \hspace{0.5cm} b. $\psi_g$ for $\Lb_g (\mathbf{v}_i, \Ns(\mathbf{v}_i))$ \tcp*{over nodes}
2. \Repeat {convergence}{ 
		\tcp{Semi-supervised}
        \For{each batch sampled from $p(j, \gamma|i, \Ds_S^l, G)$}{
		\Na Compute loss $\Ls_G(\Lambda, \Omega)$ \\
		\Nb Take a gradient step for $\Ls_G(\Lambda, \Omega)$; \\
        }
		\tcp{Supervised \& domain adversary} 
        \For{each batch sampled from $\Ds_S^l$}{
		\Na Compute $\Ls_C(\Lambda, \Phi)$	and $\Ls_D(\Lambda, \Psi)$ \\		 \Nb Take gradient steps for $\Ls_C(\Lambda, \Phi)$ and $\Ls_D(\Lambda, \Psi)$; \\
        }
		\tcp{Domain adversary} 
        \For{each batch sampled from $\Ds_S^u$ and $\Ds_T^u$}{
		\Na Compute $\Ls_D(\Lambda, \Psi)$ \\		
		\Nb Take a gradient step for $\Ls_D(\Lambda, \Psi)$;
        }
   }   
\caption{Model Training with SGD}
\label{alg:training}
\end{algorithm}

%\vspace{-2mm}

%For the training we used adadelta and trained the model in the mini-batch mode. We first sample a batch of graph context and take a gradient step to  optimize the graph context loss. Then, we sample a batch of labeled instances  from the source domain and take a gradient step to compute the loss function for the prediction of class label and domain label (i.e., source part). After that we take a batch of target instances and take a gradient step  to compute  the loss function for the prediction of domain label (i.e., target part).  

Algorithm \ref{alg:training} illustrates the training algorithm based on stochastic gradient descent (SGD). We first initialize the model parameters. The word embedding matrix $E$ is initialized with pre-trained word2vec vectors (see Subsection \ref{ssec:word_embedding}) and is kept fixed during training.\footnote{Tuning $E$ on our task by backpropagation increased the training time immensely (3 days compared to 5 hours on a Tesla GPU) without any significant performance gain.} Other parameters are initialized with small random numbers sampled from a uniform distribution \cite{Xavier10}. We use AdaDelta \cite{zeiler2012adadelta} adaptive update to update the parameters.

In each iteration, we do three kinds of gradient updates to account for the three different loss components. First, we do an epoch over all the training instances updating the parameters for the semi-supervised loss, then we do an epoch over the labeled instances in the source domain, each time updating the parameters for the supervised and the domain adversary losses. Finally, we do an epoch over the unlabeled instances in the two domains to account for the domain adversary loss.

The main challenge in adversarial training is to balance the competing components of the network. If one component becomes smarter than the other, its loss to the shared layer becomes useless, and the training fails to converge \cite{ArjovskyCB17}. Equivalently, if one component becomes weaker, its loss overwhelms that of the other, causing the training to fail. In our experiments, we observed the domain discriminator is weaker than the rest of the network. This could be due to the noisy nature of tweets, which makes the job for the domain discriminator harder. To balance the components, we would want the error signals from the discriminator to be fairly weak, also we would want the supervised loss to have more impact than the semi-supervised loss. In our experiments, the weight of the domain adversary loss $\lambda_d$ was fixed to $1e-8$, and the weight of the semi-supervised loss $\lambda_g$ was fixed to $1e-2$. Other sophisticated weighting schemes have been proposed recently \cite{ganin2016domain,ArjovskyCB17,MetzPPS16}. It would be interesting to see how our model performs using these advanced tuning methods, which we leave as a future work.

%In future 
%We follow the weighting schedule proposed by \newcite [p. 21]{Ganin:2016:DTN:2946645.2946704}, which initializes $\lambda$ to $0$, and changes it gradually to $1$ as training progresses. I.e., we start with training the task classifier first, and we gradually add the discriminator's loss.   

\subsection{Crisis Word Embedding}
\label{ssec:word_embedding}

As mentioned, we used word embeddings that are pre-trained on a crisis dataset. To train the word-embedding model, we first pre-processed tweets collected using the AIDR system \cite{imran2014aidr} during different events occurred between 2014 and 2016. In the preprocessing step, we lowercased the tweets and removed URLs, digit, time patterns, special characters, single character, username started with the $@$ symbol. After pre-processing, the resulting dataset contains about $364$ million tweets and about 3 billion words. 

There are several approaches to train word embeddings such as continuous bag-of-words (CBOW) and skip-gram models of wrod2vec \cite{mikolov2013efficient}, and Glove \cite{pennington2014glove}. For our work, we trained the CBOW  model from word2vec. While training CBOW, we filtered out words with a frequency less than or equal to 5, and we used a context window size of 5 and $k=5$ negative samples. The resulting embedding model contains about 2 million words with vector dimensions of 300.  %The crisis dataset consists of different collections of tweets collected automatically using the AIDR system \cite{imran2014aidr}.

%We initialize the embedding matrix $E$ in our network with pretrained word embeddings. 
%We trained a continuous bag-of-words (CBOW) wrod2vec \cite{mikolov2013efficient} model on a large crisis dataset 
%To design the word embeddings, we used Mikolov et
%al.~\cite{mikolov2013efficient,mikolov2013distributed} implementation of wrod2vec, which contains both continuous
%bag-of-words (CBOW) and skip-gram algorithms. 

%We designed our model
%using the CBOW approach with a size of the feature vector $300$, a
%context window size $5$, negative-sampling with a value of $k=5$ 
%and 
%\red{filtered 
%We made this model publicly available for research purpose on [LINK].} 

\section{Experimental Settings}
\label{sec:experiments}
In this section, we describe our experimental settings -- datasets used, settings of our models, compared baselines, and evaluation metrics.

%first describe the datasets we used in our experiments, then we describe the experimental setting

%, and finally we discuss the results. 

\subsection{Datasets}
\label{ssec:data}
To conduct the experiment and evaluate our system, we used two real-world Twitter datasets collected during the \textit{2015 Nepal earthquake} (NEQ) and the \textit{2013 Queensland floods} (QFL). These datasets are comprised of millions of tweets collected through the Twitter streaming API\footnote{https://dev.twitter.com/streaming/overview} using event-specific keywords/hashtags. 

To obtain the labeled examples for our task we employed paid workers from the Crowdflower\footnote{http://crowdflower.com} -- a crowdsourcing platform. The annotation consists of two classes \textit{relevant} and \textit{non-relevant}. For the annotation, we randomly sampled 11,670 and 10,033 tweets from the Nepal earthquake and the Queensland floods datasets, respectively. Given a tweet, we asked crowdsourcing workers to assign the \textit{``relevant''} label if the tweet conveys/reports information useful for crisis response such as a report of injured or dead people, some kind of infrastructure damage, urgent needs of affected people, donations requests or offers, otherwise assign the \textit{``non-relevant''} label. We split the labeled data into 60\% as training, 30\% as test and 10\% as development. Table~\ref{table:data_dist} shows the resulting datasets with class-wise distributions. 
Data preprocessing was performed by following the same steps used to train the word2vec model (Subsection \ref{ssec:word_embedding}). In all the experiments, the classification task consists of two classes: \textit{relevant} and \textit{non-relevant}.

%Table~\ref{table:data_dist} shows the class-wise distribution for the two datasets. %NEQ has a total of $11,670$ labeled tweets and QFL has a total of $10,033$ labeled tweets. We see that both datasets are fairly balanced.  

%, which shows that the NEQ dataset is fairly balanced than the QFL dataset. %\red{Say couple of sentences regarding the numbers}

\begin{table}[t]
	\centering
	%\caption{Distribution of the dataset with training, development and test split.}
	\scalebox{0.75}{
	\begin{tabular}{l|r|r|r|r|r}
		\midrule
		\multicolumn{1}{c|}{\textbf{Dataset}} & \multicolumn{1}{c|}{\textbf{Relevant}} & \multicolumn{1}{c|}{\textbf{Non-relevant}} &\multicolumn{1}{c|}{\textbf{Train}} & \multicolumn{1}{c|}{\textbf{Dev}} & \multicolumn{1}{c}{\textbf{Test}} \\ \midrule
		NEQ & 5,527 & 6,141 & 7,000 & 1,167 & 3,503 \\ \midrule
		QFL & 5,414 & 4,619 &6,019 & 1,003 & 3,011 \\ \midrule
	\end{tabular}
}
	\vspace{-0.3em}
	\caption{Distribution of labeled datasets for Nepal earthquake (NEQ) and Queensland flood (QFL).}
	\label{table:data_dist}
\end{table}

%\subsection{Experimental Settings}
%\label{ssec:exp_settings}

 %For all of our experimental settings, we considered them as binary classification task {\em relevant} \textit{vs} {\em irrelevant} to the event. 
%{For each experiment, we used the development set to optimize  hyperparameters of the model.} 

\subsection{Model Settings and Baselines}

In order to demonstrate the effectiveness of our joint learning approach, we performed a series of experiments. To understand the contribution of different network components, we performed an ablation study showing how the model performs as a semi-supervised model alone and as a domain adaptation model alone, and then we compare them with the combined model that incorporates all the components.

%A number of experiments were performed for the both \textit{semi-supervised} and \textit{domain adaptation} settings. %  using the datasets discussed in the last Section. 

\subsubsection{Settings for Semi-supervised Learning}
\label{sssec:exp_settings-semi-supervised}
As a baseline for the semi-supervised experiments, we used the self-training approach \cite{scudder1965probability}. 
%\footnote{It is also termed as label propagation in the literature}
%For the baseline experiment of semi-supervised learning we followed self-training approach \cite{scudder1965probability} \footnote{it is also termed as label propagation in the literature} to utilize the unlabelled data. 
For this purpose, we first trained a supervised model using the CNN architecture (i.e., shared components followed by the supervised part in Figure \ref{fig:cnn_graph}). The trained model was then used to automatically label the unlabeled data. Instances with a classifier confidence score $\geq0.75$ were then used to retrain a new model.

Next, we run experiments using our graph-based semi-supervised approach (i.e., shared components followed by the supervised and semi-supervised parts in Figure \ref{fig:cnn_graph}), which exploits unlabeled data. For reducing the computational cost, we randomly selected $50K$ unlabeled instances from the same domain. 
For our semi-supervised setting, one of the main goals was to understand how much labeled data is sufficient to obtain a reasonable result. Therefore, we experimented our system by incrementally adding batches of instances, such as 100, 500, 2000, 5000, and all instances from the training set. Such an understanding can help us design the model at the onset of a crisis event with sufficient amount of labeled data. 
To demonstrate that the semi-supervised approach outperforms the supervised baseline, we run supervised experiments using the same number of labeled instances. In the supervised setting, only $\mathbf{z}_c$ activations in Figure \ref{fig:cnn_graph} are used for classification.

\subsubsection{Settings for Domain Adaptation}
\label{sssec:exp_settings-domain-adaptation}
To set a baseline for the domain adaptation experiments, we train a CNN model (i.e., shared components followed by the supervised part in Figure \ref{fig:cnn_graph}) on one event (source) and test it on another event (target). We call this as \emph{transfer baseline}. 

To assess the performance of our domain adaptation technique alone, we exclude the semi-supervised component from the network. We train and evaluate models with this network configuration using different source and target domains. 

Finally, we integrate all the components of the network %%(the complete system depicted in Figure as 
as shown in Figure \ref{fig:cnn_graph} and run domain adaptation experiments using different source and target domains. In all our domain adaptation experiments, we only use unlabeled instances from the target domain. In domain adaption literature, this is known as \emph{unsupervised} adaptation. 

%Upon obtaining a promising results, then we integrated graph-based approach with adversarial based domain adaptation for unsupervised domain adaptation (i.e., considering only unlabeled target data). For the experiments of adversarial training with graph based approach we used the whole system presented in Figure \ref{fig:cnn_graph}.   

\subsubsection{Training Settings}
\label{sssec:modelsetting}

We use 100, 150, and 200 filters each having the window size of 2, 3, and 4, respectively, and pooling length of 2, 3, and 4, respectively. We do not tune these hyperparameters in any experimental setting since the goal was to have an end-to-end comparison with the same hyperparameter setting and understand whether our approach can outperform the baselines or not. Furthermore, we do not filter out any vocabulary item in any  settings. %Moreover, we do not fine-tune the word embeddings on the classification task.

As mentioned before in Subsection \ref{subsec:training}, we used AdaDelta \cite{zeiler2012adadelta} to update the model parameters in each SGD step. The learning rate was set to $0.1$ when optimizing on the classification loss and to $0.001$ when optimizing on the semi-supervised loss. The learning rate for domain adversarial training was set to $1.0$. The maximum number of epochs was set to 200, and dropout rate of $0.02$ was used to avoid overfitting~\cite{srivastava2014dropout}. We used validation-based \emph{early stopping} using the F-measure %We did \emph{early stopping} based on the F-measure on the validation set 
with a patience of 25, i.e., we stop training if the score does not increase for 25 consecutive epochs.

\subsubsection{Evaluation Metrics}
\label{sssec:evaluation}
To measure the performance of the trained models using different approaches described above, we use weighted average precision, recall, F-measure, and Area Under ROC-Curve (AUC), which are standard evaluation measures in the NLP and machine learning communities. 
%We have measured the performance of the system using the weighted average precision, recall, F-measure and Area Under a Curve (AUC). These are commonly used evaluation measure in NLP and machine learning community. 
%For the experiment, in all experimental settings, we usually used the development set for the perameter learning and used test set for final performance measure. 
The rationale behind choosing the weighted metric is that it takes into account the class imbalance problem. %In Section \ref{ssec:results_and_discussion}, we only present the results on the test sets.
% (see Table \ref{table:data_dist}). 

\section{Results and Discussion}
\label{sec:results}
%\subsection{Results and Discussion}
%\label{ssec:results_and_discussion}

In  this section, we present the experimental results and discuss our main findings. 

\subsection{Semi-supervised Learning}
\label{sssec:results-semi-supervised}
In Table \ref{table:results_semisupervised}, we present the results obtained from the supervised, self-training based semi-supervised, and our graph-based semi-supervised experiments for the both datasets. It can be clearly observed that the graph-based semi-supervised approach outperforms the two baselines -- supervised and self-training based semi-supervised. Specifically, the graph-based approach shows 4\% to 13\% absolute improvements in terms of F1 scores for the Nepal and Queensland datasets, respectively.

\begin{table}[t!]
\centering
\scalebox{0.70}{
\begin{tabular}{l|r|r|r|r}
\midrule
\multicolumn{1}{c|}{\textbf{Experiments}} & \multicolumn{1}{c|}{\textbf{AUC}} & \multicolumn{1}{c|}{\textbf{P}} & \multicolumn{1}{c|}{\textbf{R}} & \multicolumn{1}{c}{\textbf{F1}} \\ \midrule
\multicolumn{5}{c}{\sc{Nepal Earthquake}} \\ \midrule
\textbf{Supervised} & 61.22 & 62.42 & 62.31 & 60.89 \\ \midrule
\textbf{Semi-supervised (Self-training)} & 61.15 & 61.53 & 61.53 & 61.26 \\ \midrule
\textbf{Semi-supervised (Graph-based)} & 64.81 & 64.58 & 64.63 & 65.11 \\ \midrule
\multicolumn{5}{c}{\sc{Queensland Floods}} \\ \midrule
\textbf{Supervised} & 80.14 & 80.08 & 80.16 & 80.16 \\ \midrule
\textbf{Semi-supervised (Self-training)} & 81.04 & 80.78 & 80.84 & 81.08 \\ \midrule
\textbf{Semi-supervised (Graph-based)} & 92.20 & 92.60 & 94.49 & 93.54 \\ \midrule
\end{tabular}
}
\caption{Results using supervised, self-training, and graph-based semi-supervised approaches in terms of Weighted average AUC, precision (P), recall (R) and F-measure (F1).}
%\red{We need CNN-Graph here}
\label{table:results_semisupervised}
\end{table}

To determine how the semi-supervised approach performs in the early hours of an event when only fewer labeled instances are available, we mimic a batch-wise (not to be confused with minibatch in SGD) learning setting. In Table \ref{table:results_learning_curve}, we present the results using different batch sizes -- 100, 500, 1,000, 2,000, and all labels. 

From the results, we observe that models' performance improve as we include more labeled data --- from $43.63$ to $60.89$ for NEQ and from $48.97$ to $80.16$ for QFL in the case of labeled only (\textbf{\textit{L}}). When we compare supervised vs. semi-supervised (\textit{\textbf{L}} vs. \textit{\textbf{L+U}}), we observe significant improvements in F1 scores for the semi-supervised model for all batches over the two datasets. As we include unlabeled instances with labeled instances from the same event, performance significantly improves in each experimental setting giving $5\%$ to $26\%$ absolute improvements over the supervised models. These improvements demonstrate the effectiveness of our approach. We also notice that our semi-supervised approach can perform above $90\%$ depending on the event. Specifically, major improvements are observed from batch size 100 to 1,000, however, after that the performance improvements are comparatively minor. The results obtained using batch sizes 500 and 1,000 are reasonably in the acceptable range when labeled and unlabeled instances are combined (i.e., \textit{L+50kU} for Nepal and \textit{L+$\sim$21kU} for Queensland), which is also a reasonable number of training examples to obtain at the onset of an event.

\begin{table}[t!]
\centering
	\scalebox{0.75}{
\begin{tabular}{l|r|r|r|r|r}
\hline
\multicolumn{1}{c|}{\textbf{Exp.}} & \multicolumn{1}{c|}{\textbf{100}} & \multicolumn{1}{c|}{\textbf{500}} & \multicolumn{1}{c|}{\textbf{1000}} & \multicolumn{1}{c|}{\textbf{2000}} & \multicolumn{1}{c}{\textbf{All L}} \\ \midrule
\multicolumn{6}{c}{\sc{Nepal Earthquake}} \\ \midrule
\textbf{L} & 43.63 & 52.89 & 56.37 & 60.11 & 60.89 \\ \midrule
\textbf{L+50kU} & 52.32 & 59.95 & 61.89 & 64.05 & 65.11 \\ \midrule
\multicolumn{6}{c}{\sc{Queensland Flood}} \\ \midrule
\textbf{L} & 48.97 & 76.62 & 80.62 & 79.16 & 80.16 \\ \midrule
\textbf{L+$\sim$21kU} & 75.08 & 85.54 & 89.08 & 91.54 & 93.54 \\ \midrule
\end{tabular}
}
\caption{Weighted average F-measure for the graph-based semi-supervised settings using different batch sizes. \textit{L} refers to labeled data, \textit{U} refers to unlabeled data, \textit{All L} refers to all labeled instances for that particular dataset.}
	\label{table:results_learning_curve}
\end{table}

\subsection{Domain Adaptation}
\label{sssec:results-domain-adaptation}
In Table \ref{table:results_domain_adaptation}, we present domain adaptation results. The first block shows event-specific (i.e., train and test on the same event) results for the supervised CNN model. These results set the upper bound for our domain adaptation methods. The \emph{transfer} baselines are shown in the next block, where we train a CNN model in one domain and test it on a different domain. Then, the third block shows the results for the domain adversarial approach without the semi-supervised loss. These results show the importance of domain adversarial component. After that, the fourth block presents the performance of the model trained with graph embedding without domain adaptation to show the importance of semi-supervised learning. The final block present the results for the complete model that includes all the loss components.

\begin{table}[t!]
\centering
\scalebox{0.70}{
\begin{tabular}{l|l|r|r|r|r}
\toprule
\multicolumn{1}{c|}{\textbf{Source}} & \multicolumn{1}{c|}{\textbf{Target}} & \multicolumn{1}{c|}{\textbf{AUC}} & \multicolumn{1}{c|}{\textbf{P}} & \multicolumn{1}{c|}{\textbf{R}} & \multicolumn{1}{c}{\textbf{F1}} \\ 
\midrule
\multicolumn{6}{c}{\sc In-domain supervised model} \\
\midrule
{\textbf{Nepal}} & \textbf{Nepal} & 61.22 & 62.42 & 62.31 & 60.89 \\  \midrule
{\textbf{Queensland}} & \textbf{Queensland} & 80.14 & 80.08 & 80.16 & 80.16 \\ 
\bottomrule
\multicolumn{6}{c}{\sc Transfer Baselines} \\
\midrule
{\textbf{Nepal}} & \textbf{Queensland} & 58.99 & 59.62 & 60.03 & 59.10 \\ 
 \midrule
{\textbf{Queensland}} & \textbf{Nepal} & 54.86 & 56.00 & 56.21 & 53.63 \\ 
\bottomrule
\multicolumn{6}{c}{\sc Domain Adversarial} \\
\midrule
\textbf{Nepal} & \textbf{Queensland} & 60.15 & 60.62 & 60.71 & 60.94 \\ \midrule
\textbf{Queensland} & \textbf{Nepal} & 57.63 & 58.05 & 58.05 & 57.79 \\ 
\bottomrule
\multicolumn{6}{c}{\sc Graph Embedding without Domain Adversarial} \\
\midrule
\textbf{Nepal} & \textbf{Queensland} & 60.38 & 60.86 & 60.22 & 60.54 \\ \midrule
\textbf{Queensland} & \textbf{Nepal} & 54.60 & 54.58 & 55.00 & 54.79 \\ 
\bottomrule
\multicolumn{6}{c}{\sc Graph Embedding with Domain Adversarial} \\
\midrule
\textbf{Nepal} & \textbf{Queensland} & 66.49 & 67.48 & 65.90 & 65.92 \\ \midrule
\textbf{Queensland} & \textbf{Nepal} & 58.81 & 58.63 & 59.00 & 59.05 \\ \bottomrule
\end{tabular}
}
\caption{Domain adaptation experimental results. Weighted average AUC, precision (P), recall (R) and F-measure (F1).}
\label{table:results_domain_adaptation}
\end{table}

The results with domain adversarial training show improvements across both events -- from $1.8\%$ to $4.1\%$ absolute gains in F1. These results attest that adversarial training is an effective approach to induce domain invariant features in the internal representation as shown previously by \citet{ganin2016domain}.

Finally, when we do both semi-supervised learning and unsupervised domain adaptation, we get further improvements in F1 scores ranging  from 5\% to 7\% absolute gains. From these improvements, we can conclude that domain adaptation with adversarial training along with graph-based semi-supervised learning is an effective method to leverage unlabeled and labeled data from a different domain.

Note that for our domain adaptation methods, we only use unlabeled data from the target domain. Hence, we foresee future improvements of this approach by utilizing a small amount of target domain labeled data. 

%In order to understand the feature representation after domain adaptation, we use t-SNE \cite{maaten2008visualizing} plot to visualize the representation layer  vectors ($\mathbf{z}$) for source and target instances. Figure \ref{fig:domain-adaptation} shows the scatter plots after the 2-D projection.  For the purpose of visualization, we randomly selected 1000 instances for each case, i.e., with and without domain adaptation. The red color represents the source and blue color represents the target training instances. We can see that the plots look as expected after training  with the domain discriminator. The domain adapted model mixes the blue and the red examples more compactly, as the adversarial part of the network pushes for learning shared abstract features that are domain-insensitive. 

%As presented in Figure \ref{fig:domain-adaptation} with domain adaptation there is a higher overlap between source and target domain distributions compared to the without domain adaptation. 

%\red{Need to write about the figure: red colour represent source and blue represent target instances}

\section{Related Work}
\label{sec:related_works}
%self-training, 

Two lines of research are directly related to our work: \Ni semi-supervised learning and \Nii domain adaptation. 
%\subsection{Semi-supervised Learning}
Several models have been proposed for semi-supervised learning. 
%The earliest approaches are self-training \cite{scudder1965probability} and co-training \cite{mitchell1999role}.
The earliest approach is self-training \cite{scudder1965probability}, in which a trained model is first used to label unlabeled data instances followed by the model re-training with the most confident predicted labeled instances. 
%co-training, 
The co-training \cite{mitchell1999role} approach assumes that features can be split into two sets and each subset is then used to train a classifier with an assumption that the two sets are conditionally independent. 
Then each classifier classifies the unlabeled data, and then most confident data instances are used to re-train the other classifier, this process repeats multiple times. 
%transductive support vector machines, 
%TSVMs \cite{vapnik1998statistical} is an extension of standard support vector machines designed with additional regularization parameter to use unlabeled data. 
%and graph-based methods. 

In the graph-based semi-supervised approach, nodes in a graph represent labeled and unlabeled instances and edge weights represent the similarity between them. The structural information encoded in the graph is then used to regularize a model \cite{zhu05survey}. 
There are two paradigms in semi-supervised  learning: 1) inductive -- learning a function with which predictions can be made on unobserved instances, 2) transductive -- no explicit function is learned and predictions can only be made on observed instances. 
As mentioned before, inductive semi-supervised learning is preferable over the transductive approach since it avoids building the graph each time it needs to infer the labels for the unlabeled instances. 

In our work, we use a graph-based inductive deep learning approach proposed by \citet{yang2016revisiting} to learn features in a deep learning model by predicting contextual (i.e., neighboring) nodes in the graph. However, our approach is different from \citet{yang2016revisiting} in several ways. First, we construct the graph by computing the distance between tweets based on word embeddings. Second, instead of using count-based features, we use a convolutional neural network (CNN) to compose high-level features from the distributed representation of the words in a tweet. Finally, for context prediction, instead of performing a random walk, we select nodes based on their similarity in the graph. Similar similarity-based graph has shown impressive results in learning sentence representations \cite{saha-17}.  

%%%%%%%%%% Domain adaptation :related work
%\cite{blitzer2006domain}
%\subsection{Domain Adaptation}
%In the last decade there has been significant amount work for domain adaptation. 
In the literature, the proposed approaches for domain adaptation include supervised, semi-supervised and unsupervised. It also varies from linear kernelized approach \cite{blitzer2006domain} to non-linear deep neural network techniques \cite{glorot2011domain,ganin2016domain}. One direction of research is to focus on feature space distribution matching by reweighting the samples from the source domain \cite{gong2013connecting} to map source into target. The overall idea is to learn a good feature representation that is invariant across domains. 
In the deep learning paradigm, Glorot et al. \cite{glorot2011domain} used Stacked Denoising Auto-Encoders (SDAs) for domain adaptation. SDAs learn a robust feature representation, which is artificially corrupted with small Gaussian noise. %Chen et al. in \cite{chen2012marginalized} used marginal SDAs, which has advantages over SDAs \cite{vincent2008extracting} in terms of computational cost. 
%In \cite{tzeng2014deep}, Tzeng et al. proposed a CNN architecture with a domain adaptation layer to learn a feature representation that is domain invariant.
\emph{Adversarial training} of neural networks has shown big impact recently, especially in areas such as computer vision, where generative unsupervised models have proved capable of synthesizing new images \cite{Goodfellow_14_GAN,RadfordMC15,MakhzaniSJG15}. 
\citet{ganin2016domain} proposed domain adversarial neural networks (DANN) to learn discriminative but at the same time domain-invariant representations, with domain adaptation as a target. We extend this work by combining with semi-supervised graph embedding for unsupervised domain adaptation. 

%\textcolor{red}{TODO} 
In a recent work, \citet{kipf2016semi} present CNN applied directly on graph-structured datasets - citation networks and on a knowledge graph dataset. Their study demonstrate that graph convolution network for semi-supervised classification performs better compared to other graph based approaches. 
%Semi-supervised classification with graph convolutional networks \cite{kipf2016semi}

\section{Conclusions}
\label{sec:conclusions}
%\vspace{-2mm}
In this paper, we presented a deep learning framework that performs domain adaptation with adversarial training and graph-based semi-supervised learning to leverage labeled and unlabeled data from related events. We use a convolutional neural network to compose high-level representation from the input, which is then passed to three components that perform supervised training, semi-supervised learning and domain adversarial training. 
%The first component induces label information into the network using a supervised loss for predicting the class labels. The second component induces structural similarity between training instances using a graph-based semi-supervised loss. We constructed the similarity graph using a k-nearest neighbor approach that exploits distributed representations of the tweets. The third component induces domain invariance into the learned representation using a domain adversary loss. 
For domain adaptation, we considered a scenario, where we have only unlabeled data in the target event. 
Our evaluation on two crisis-related tweet datasets demonstrates that by combining domain adversarial training with semi-supervised learning, our model gives significant improvements over their respective baselines. We have also presented results of batch-wise incremental training of the graph-based semi-supervised approach and show approximation regarding the number of labeled examples required to get an acceptable performance at the onset of an event. 
%There are several interesting future research directions of this work. First, we would like to experiment with the domain adaptation scenario where there are few labeled data in the target domain. Second, we would like to explore zero- or one-shot learning, where the goal is to design models, which would be capable of learning a new label from one or zero example. 

%\section*{Acknowledgments}

%The acknowledgments should go immediately before the references.  Do not number the acknowledgments section ({\em i.e.}, use \verb|\section*| instead of \verb|\section|). Do not include this section when submitting your paper for review.

%\small
{
\bibliographystyle{acl}
\balance
%\bibliography{acl2018}
\bibliography{./bib/biblist,./bib/biblist_ml,./bib/bibliography,./bib/ijcai17,./bib/biblist_domain_adaptation}
}
%\appendix

\end{document}